%% file: main.tex
\documentclass{article}

\usepackage{microtype}
\usepackage{graphicx}
\usepackage{subcaption}
\usepackage{booktabs} 

\usepackage{hyperref}

\usepackage[preprint]{icml2026}

\usepackage{amsmath}
\usepackage{amssymb}
\usepackage{mathtools}
\usepackage{amsthm}

\usepackage[capitalize,noabbrev]{cleveref}

\usepackage[table,xcdraw]{xcolor}
\usepackage{xspace}
\usepackage{tcolorbox}
\tcbuselibrary{skins}
\usepackage{pifont}

\theoremstyle{plain}

\theoremstyle{definition}

\theoremstyle{remark}

\newcommand{\ie}{\textit{i}.\textit{e}.}
\newcommand{\eg}{\textit{e}.\textit{g}.}

\newcommand{\etc}{\textit{etc}.}

\newcommand{\model}{\texttt{DualSpeed}\xspace}

\usepackage[textsize=tiny]{todonotes}

\icmltitlerunning{Fast-Slow Efficient Training for Multimodal Large Language Models via Visual Token Pruning}

\begin{document}

\twocolumn[
  \icmltitle{Fast-Slow Efficient Training for Multimodal Large Language Models\\ via Visual Token Pruning}



  \icmlsetsymbol{equal}{*}

  \begin{icmlauthorlist}
    \icmlauthor{Dingkun Zhang}{1}
    \icmlauthor{Shuhan Qi}{1}
    \icmlauthor{Yulin Wu}{1}
    \icmlauthor{Xinyu Xiao}{1}
    \icmlauthor{Xuan Wang}{1}
    \icmlauthor{Long Chen}{2}
  \end{icmlauthorlist}

  \icmlaffiliation{1}{Harbin Institute of Technology, Shenzhen}
  \icmlaffiliation{2}{Hong Kong University of Science and Technology}

  \icmlcorrespondingauthor{Shuhan Qi}{shuhanqi@cs.hitsz.edu.cn}


  \vskip 0.3in
]



\printAffiliationsAndNotice{}  

\begin{abstract}
Multimodal Large Language Models (MLLMs) suffer from severe training inefficiency issue, which is associated with their massive model sizes and visual token numbers.
Existing efforts in efficient training focus on reducing model sizes or trainable parameters.
Inspired by the success of Visual Token Pruning (VTP) in improving inference efficiency, we are exploring another substantial research direction for efficient training by reducing visual tokens.
However, applying VTP at the training stage results in a \textit{training-inference mismatch}: pruning-trained models perform poorly when inferring on non-pruned full visual token sequences.
To close this gap, we propose \model, a fast-slow framework for efficient training of MLLMs.
The fast-mode is the primary mode, which incorporates existing VTP methods as plugins to reduce visual tokens, along with a mode isolator to isolate the model's behaviors.
The slow-mode is the auxiliary mode, where the model is trained on full visual sequences to retain training-inference consistency.
To boost its training, it further leverages self-distillation to learn from the sufficiently trained fast-mode.
Together, \model can achieve both training efficiency and non-degraded performance.
Experiments show \model accelerates the training of LLaVA-1.5 by 2.1$\times$ and LLaVA-NeXT by 4.0$\times$, retaining over 99$\%$ performance.
Code: \url{https://github.com/dingkun-zhang/DualSpeed}.
\end{abstract}

\input{sections/introduction}

\input{sections/related_work}

\input{sections/method}

\input{sections/experiments}

\section{Conclusion}
In this paper, we introduce a novel framework \model to accelerate MLLM training.
It comprises a fast-mode and a slow-mode, to incorporate existing VTP methods as plugins to reduce visual tokens and to retain training-inference consistency, respectively.
Through this fast-slow manner, \model can achieve both training acceleration and non-degraded performance.
Extensive experiments on diverse benchmarks show that the proposed framework achieves significant training speedup with nearly lossless performance across different MLLMs and resolutions.

\section*{Impact Statement}
This work enables visual token pruning for efficient Multimodal Large Language Model (MLLM) training, reducing costs while preserving performance.
For social impact, faster, cheaper MLLM development accelerates applications in healthcare diagnostics, education, and accessibility tools.
However, over-pruning may degrade model robustness, threatening reliability in high-risk scenarios.
Moreover, reduced training barriers could enable malicious actors to deploy harmful multimodal models at scale.
Mitigating these requires strict validation and ethical guidelines for responsible MLLM deployment.

\bibliography{main}
\bibliographystyle{icml2026}

\newpage
\appendix
\onecolumn
\input{sections/appendix}

\end{document}

%% file: sections/introduction.tex
\section{Introduction}\label{sec:intro}

\begin{figure}[!t]
\vskip 0.2in
\centering
\includegraphics[width=\columnwidth]{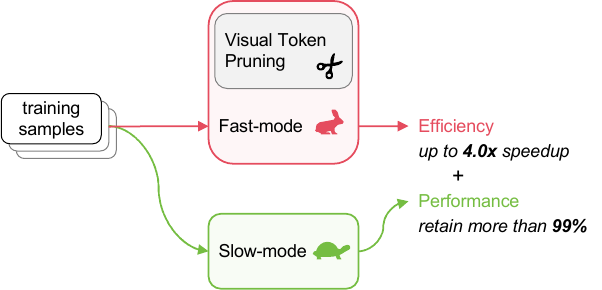}
\caption{\textbf{\model significantly accelerates MLLM training while retaining performance.} It achieves up to 4.0$\times$ training speedup, with nearly lossless performance.}
\vskip -0.1in
\label{fig:intro}
\end{figure}

Multimodal Large Language Models (MLLMs)~\cite{llava,llava_1.5,llava_next,instructblip,internvl,qwen_vl,qwen2_vl,qwen2.5_vl} have achieved remarkable breakthroughs in bridging visual and linguistic modalities, enabling superior performance on a wide range of cross-modal tasks such as image captioning~\cite{captioning}, visual question answering~\cite{vqa,vqav2}, visual grounding~\cite{flickr30k_entities}, and multimodal reasoning~\cite{sqa,gqa}.
The core of their success lies in effective visual-language pretraining~\cite{vila}, where visual content is converted into visual tokens and processed along with textual tokens by the Large Language Models (LLMs)~\cite{llama,llama2,gpt4}.
However, the rapid growth of model size, visual resolution, \etc, has brought a critical bottleneck in training efficiency.
This efficiency bottleneck severely restricts the scalability~\cite{mera} of MLLMs, making large-scale pretraining resource-intensive and thus prohibitive for many research and industrial applications.

Generally, the training efficiency of MLLMs is associated with two core factors: the size of trainable parameters and the number of training tokens.
The two factors collectively determine the computational complexity, memory, and time consumption.
Existing efforts towards efficient training mainly focus on reducing the total model size or trainable parameters, \eg, model compression~\cite{llava_kd,q_vlm,laptop_diff}, parameter-efficient tuning~\cite{llava_steering,tuning_layernorm_mllm,peft_mllm}, and vision encoder grafting~\cite{encoder_grafting}.
On the other side, research on reducing the number of tokens during training is orthogonal to these fields and has substantial potential.
From the token-centric perspective, a primary driver of the heavy training costs is the massive number of visual tokens, often hundreds or thousands.

\begin{figure}[!t]
\vskip 0.2in
\centering
\includegraphics[width=\columnwidth]{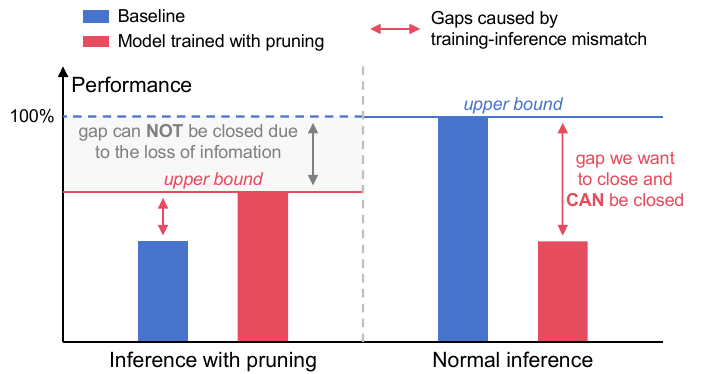}
\caption{\textbf{Illustration of the performance gaps caused by training-inference mismatch.} The performance upper bound of pruning inference is theoretically lower than normal inference, as pruning leads to the loss of visual information. Therefore, the possible path to approaching 100\% performance is to address the training-inference mismatch.}
\label{fig:close_gap}
\end{figure}

To address the issue of excessive visual tokens, Visual Token Pruning (VTP)~\cite{fastv,fastervlm,divprune,cdpruner,mmtok} is proposed to reduce the redundant visual tokens.
The core insight from this field is that not all visual tokens are essential—many tokens correspond to redundant or low-informative regions, \eg, homogeneous backgrounds, repetitive textures, which do not provide additional value for multimodal understanding, and thus can be dropped efficiently.
However, despite the principle of it being universal for both inference time and training time, VTP suffers from severe training-inference mismatch when applied in training.
Here, we define the \textbf{training-inference mismatch} as: pruning-trained models perform poorly when inference on non-pruned full visual sequences.
This mismatch may originate from discrepancies in sequence length, information density, spatial structure, \etc, between pruned and full visual sequences.
As illustrated in Figure~\ref{fig:close_gap}, when trained with pruning, the model struggles to perform effective inference on non-pruned full visual sequences.
This performance gap is further empirically measured in Section~\ref{sec:train_infer_gap}.

To integrate VTP into the training pipeline without causing training-inference mismatch, we propose the dual-speed mode training via visual token pruning (\model).
It comprises a fast-mode and a slow-mode, which switch randomly during training.
The fast-mode serves as the primary training mode, which incorporates existing VTP methods as plugins to reduce visual tokens.
To enable the adaptation to both pruned and full visual inputs, a learnable mode isolator is introduced.
It is concatenated with pruned visual sequences as a prefix to prompt the model to activate a specific perception pattern, and is disabled when processing full visual sequences to switch the model to another pattern.
The slow-mode acts as an auxiliary mode, training on full visual sequences to retain training-inference consistency.
Considering that the slow-mode is switched to less frequently due to the dominance of fast-mode, we incorporate self-distillation to enhance its learning effectiveness.
The sufficiently trained fast-mode acts as a teacher to guide the slow-mode, with teacher and student sharing the same model parameters but differing in input sequences.
This design ensures that the slow-mode can learn effectively from the fast-mode while retaining the capability to process full visual inputs.
By utilizing these two modes, the \model framework achieves efficient training via VTP while retaining training-inference consistency.

We highlight the main contributions of this paper below:
\begin{itemize}
\item We introduce visual token pruning to accelerate the training of MLLMs, and identify a key challenge of training-inference mismatch.
\item We propose \model, which leverages collaborative fast-slow modes, to enable efficient training while addressing the training-inference mismatch.
\item Our framework achieves 2.1$\times$ training speedup on LLaVA, and 4.0 $\times$ training speedup on LLaVA-NeXT, retraining over 99$\%$ performance on various vision-language benchmarks.
\end{itemize}

%% file: sections/related_work.tex
\section{Related Work}

\textbf{Multimodal Large Language Models.}
Multimodal Large Language Models (MLLMs)~\cite{llava,llava_1.5,llava_next,qwen_vl,instructblip,internvl} typically consist of three core components: a vision encoder~\cite{clip,vit}, a multimodal projector~\cite{llava,llava_1.5}, and an LLM~\cite{llama,llama2}.
The visual encoder, often instantiated by Vision Transformers (ViTs)~\cite{vit} or their variants, converts raw visual inputs, \eg, images, video frames, into visual tokens.
For instance, LLaVA~\cite{llava,llava_1.5} adopts CLIP-ViT-L/14~\cite{clip_vit_l_14} to generate 576 visual tokens per 336$\times$336 resolution image, while higher-resolution inputs further escalate token numbers to thousands.
These visual tokens are then projected into the text space by the multimodal projector, enabling the LLM backbone to process multimodal inputs uniformly.
However, the massive number of visual tokens, coupled with the quadratic computational complexity of self-attention in LLMs, becomes a critical bottleneck for training efficiency.

\textbf{Visual Token Reduction.}
Existing visual token reduction methods for multimodal tasks can be mainly categorized into projector optimization, token merging, and token pruning approaches.
Projector optimization methods~\cite{honeybee,tokenpacker} aim to design specific multimodal projectors that can compress visual tokens, \eg, via pooling~\cite{honeybee} or multiscale aggregation~\cite{tokenpacker}.
However, their projector-specific optimization harms generalization and scalability across different architectures.
Token merging methods~\cite{tome,mustdrop} aggregate redundant tokens into compact and representative embeddings.
An inherent limitation of this category is that the aggregation operation blurs spatial locality across multiple tokens and alters the feature distribution of visual tokens.
In contrast, Visual Token Pruning (VTP) methods~\cite{fastv,fastervlm,divprune,cdpruner,mmtok} are architecture agnostic and do not alter the feature distribution of visual token.
VTP dynamically identifies and discards redundant tokens based on certain criteria, \eg, attention score~\cite{fastv,fastervlm}, diversity~\cite{divprune}, or conditional diversity~\cite{cdpruner,mmtok}.
However, existing VTP methods are inference-oriented, which overlook the training dynamics.
Directly applying VTP techniques at training time disrupts the model’s learning process, leading to severe training-inference mismatch.

%% file: sections/method.tex
\section{Method}

\begin{figure*}[!t]
\vskip 0.2in
\centering
\includegraphics[width=0.9\textwidth]{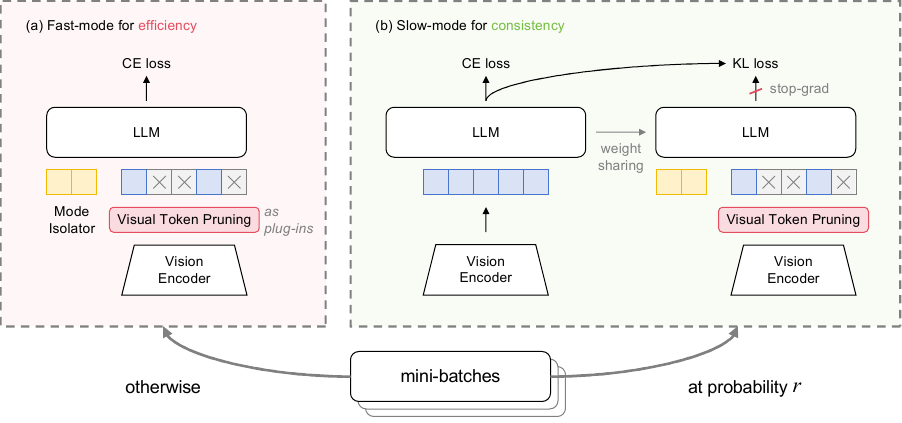}
\caption{\textbf{Overview of \model.} The framework comprises a fast-mode and a slow-mode. The two modes are randomly switched between different mini-batches, with fast-mode being the primary mode to maximize training efficiency, and slow-mode being the auxiliary mode to retain training-inference consistency. Different VTP methods are plugable for the \model framework.}
\label{fig:method}
\end{figure*}

\subsection{Visual Token Pruning}
Existing Multimodal Large Language Models (MLLMs) typically consist of three core components: a vision encoder, a multimodal projector, and an LLM.
The visual content is first encoded into tokens by the vision encoder and subsequently projected into the text space by the multimodal projector.
We denote the projected visual tokens as $E_v=\{v_1, v_2, \dots, v_n\}\in\mathbb{R}^{n\times d}$, where $n$ is the total number of visual tokens, $d$ is the hidden dimension of LLM.

Visual Token Pruning (VTP) aims to identify and reduce the redundant visual tokens according to certain criteria, \eg, attention score~\cite{fastv}, or diversity~\cite{divprune}, while preserving the most visual information.
In general, VTP optimizes a visual token subset:
\begin{equation}
\widetilde{E}_v=\textrm{VTP}(E_v,k) = \{v_1, v_2, \dots, v_k\}\in\mathbb{R}^{k\times d},
\label{eqn:vtp}
\end{equation}
where $\widetilde{E}_v\subset E_v$, and $k$ is the length of pruned visual token sequence.
If a pruning ratio $p\in(0,1)$ is given, $k$ can be computed as $k=(1-p)\cdot n$.
In practice, it is usually set to $k\ll n$, thereby significantly reducing visual tokens.
Despite its proven effectiveness in inference, when applied in training, VTP leads to severe training-inference mismatch, which may stem from the discrepancies in sequence length, information density, \etc\xspace(discussed in Appendix~\ref{app_sec:cause_of_mismatch}).

\subsection{Dual-Speed Mode Training (\model)}
To effectively integrate VTP into the training stage without causing training-inference mismatch, we propose \model.
It leverages a fast-slow manner, which comprises two distinctive modes.
The \textit{fast-mode} is the primary mode, where different VTP methods can be seamlessly plugged in, and then the model can be efficiently trained on pruned visual sequences.
The \textit{slow-mode} is the auxiliary mode to mitigate the training-inference mismatch, where the model learns on non-pruned full visual sequences.
Formally, the fast and slow modes are switched as:
\begin{equation}
\begin{split}
\text{mode}=
\begin{cases}
\text{Slow-Mode}, &\text{at probability $r$ per batch},\\
\text{Fast-Mode}, &\text{otherwise},
\end{cases}
\end{split}
\end{equation}
where $r\in(0,1)$ is the slow-mode activation probability.

\subsubsection{Fast-Mode}
In fast-mode, the model is efficiently trained on pruned visual sequences.
Given a pruning ratio $p$, it first leverages a certain VTP method to obtain the pruned visual token sequence $\widetilde{E}_v$ according to Equation~\ref{eqn:vtp}.
Thereby, the model can efficiently learns from shorter visual sequences while keeping most of the visual information.

Notably, the model should eventually also be capable of perceiving full visual sequences that are longer, more sparse, but contain greater amounts of information.
Therefore, we propose to isolate the model's perception patterns to accommodate pruned or full visual sequences at the input end.

\textbf{Mode Isolator.}
A mode isolator $P\in\mathbb{R}^{l\times d}$ is realized as a learnable soft prompt~\cite{ptuning} of length $l$.
It is concatenated as a prefix with the pruned visual sequence $\widetilde{E}_v$ to explicitly prompt the LLM to activate a distinctive perception pattern.
Conversely, for non-pruned full visual sequences, $P$ is not employed, so the LLM can exhibit another perception pattern.
With the mode isolator, fast-mode learns from the visual sequence:
\begin{equation}
\widehat{E}_v=P\oplus\widetilde{E}_v\in\mathbb{R}^{(l+k)\times d},
\end{equation}
where $\oplus$ is the concatenation operation, and the length $l$ of $P$ is short enough for efficient training.

\textbf{Training Objective of Fast-Mode.}
Fast-mode adopts the originally adopted Cross-Entropy loss.
We denote the original training objective of MLLMs as:
\begin{equation}
\mathcal{L}_{\mathrm{CE}}(E_v)=-\frac{1}{m}\sum_{t=1}^{m}\log p_\theta\left(y_t\mid E_v,y_{1:t-1}\right),
\end{equation}
where $p_\theta$ denotes the conditional probability distribution predicted by the model $\theta$, and $y=\{y_1, y_2, \dots, y_m\}$ is the textual token sequence.
Therefore, the training objective of fast-mode is computed as:
\begin{equation}
\mathcal{L}_{\mathrm{fast}}=\mathcal{L}_{\mathrm{CE}}(\widehat{E}_v).
\label{eqn:loss_fast}
\end{equation}

\subsubsection{Slow-Mode}
Slow-mode is the auxiliary mode, where the model is trained on non-pruned visual sequences to enforce training-inference consistency.
Firstly, slow-mode adopts the original training objective $\mathcal{L}_{\mathrm{CE}}(E_v)$, computed on non-pruned sequences.
However, in practice, the slow-mode may not be sufficiently trained since most batches are learned in the primary fast-mode for training efficiency.

\textbf{Self-Distillation.}
To boost the learning of slow-mode, we introduce self-distillation to facilitate knowledge reuse between fast and slow modes.
We utilize the sufficiently trained fast-mode as the teacher to guide the learning of slow-mode student.
The distillation loss $\mathcal{L}_{\mathrm{distill}}$ is computed as the Kullback–Leibler (KL) divergence between the output logits of the teacher and student:
\begin{equation}
\begin{split}
\mathcal{L}_{\mathrm{distill}}&(\widehat{E}_v,E_v)=\sum_{t=1}^{m}\mathrm{KL}\left(p_t^{\mathrm{tea}}\mid\mid p_t^{\mathrm{stu}}\right),\\
p_t^{\mathrm{tea}}&=\mathrm{Softmax}[z_\theta(y_t\mid\widehat{E}_v,y_{1:t-1})/\tau],\\
p_t^{\mathrm{stu}}&=\mathrm{Softmax}[z_\theta(y_t\mid E_v,y_{1:t-1})/\tau],
\end{split}
\end{equation}
where $p_t^{\mathrm{tea}}$ and $p_t^{\mathrm{stu}}$ are the temperature-scaled logits distributions from teacher and student, respectively, $z_\theta$ denotes the LLM logtis, and $\tau$ is the distillation temperature.
The teacher and student share the same model parameters, with the only difference being their inputs.
This additional distillation loss can be efficiently computed, since the teacher forward passes on the pruned sequence and does not compute gradients or perform backpropagation.

\textbf{Training Objective of Slow-Mode.}
The final training objective of slow-mode is computed as:
\begin{equation}
\mathcal{L}_{\mathrm{slow}}=\mathcal{L}_{\mathrm{CE}}(E_v)+\mathcal{L}_{\mathrm{distill}}(\widehat{E}_v,E_v).
\label{eqn:loss_slow}
\end{equation}
Despite the limited capability of the teacher (compared to a model trained without pruning), the Cross-Entropy loss $\mathcal{L}_{\mathrm{CE}}(E_v)$ allows the student to learn with full visual information, and eventually surpass the teacher.

\subsection{Overall Framework}

\textbf{Training Stage.}
In summary, fast-mode trains on $\widehat{E}_v$, \ie, pruned visual sequence prefixed with mode isolator, and slow-mode trains on full visual sequence $E_v$ with additional self-distillation loss.
The overview of \model is illustrated in Figure~\ref{fig:method}.
The overall training objective is:
\begin{equation}
\begin{split}
\mathcal{L}&=(1-\mathbb{I}(r))\cdot\mathcal{L}_{\mathrm{fast}}+\mathbb{I}(r)\cdot\mathcal{L}_{\mathrm{slow}},\\
\mathbb{I}(r)&=
\begin{cases}
1,\ \mathrm{at\ probability}\ r\ \mathrm{per\ batch},\\
0,\ \mathrm{otherwise},
\end{cases}
\end{split}
\end{equation}
where $\mathbb{I}$ is the indicator function, and $\mathcal{L}_{\mathrm{fast}}$ and $\mathcal{L}_{\mathrm{slow}}$ are given in Equation~\ref{eqn:loss_fast} and Equation~\ref{eqn:loss_slow}, respectively.
In practice, the training-inference mismatch occurs only during the training of the LLM (further discussion and empirical evidence are in Appendix~\ref{app_sec:which_part_mismatch}).
Therefore, to further enhance efficiency, we activate solely the fast-mode in training phases where the LLM is kept frozen\footnote{For instance, the pretraining phase of the LLaVA series keeps the LLM frozen.}.

\textbf{Inference Stage.}
The inference configurations are consistent with the models trained without \model, \ie, perform inference on non-pruned full visual sequences $E_v$, to achieve uncompromised performance.

%% file: sections/experiments.tex
\section{Experiments}

\begin{figure*}[!t]
\vskip 0.2in
\centering
\begin{subfigure}[!t]{\columnwidth}
\includegraphics[width=\columnwidth]{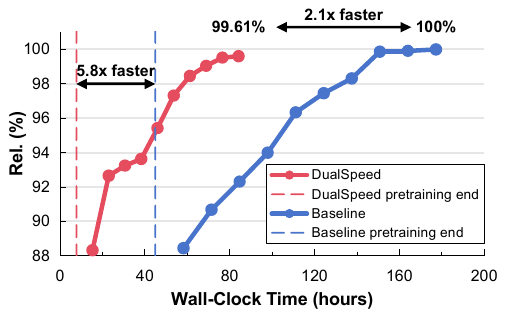}
\caption{Results on LLaVA-1.5-7B}
\label{fig:result_llava}
\end{subfigure}
\begin{subfigure}[!t]{\columnwidth}
\includegraphics[width=\columnwidth]{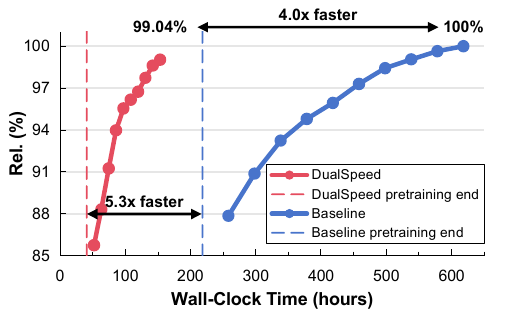}
\caption{Results on LLaVA-NeXT-7B\textsuperscript{$\dagger$}}
\label{fig:result_llava_next}
\end{subfigure}
\caption{\textbf{Comparison of training time and performance between \model and the baseline.} We measure the performance of 10 checkpoints uniformly saved during the SFT phase. Performance in the pretraining phase is not measured since the model in this phase can not follow the common instructions of many benchmarks. The wall-clock time is recorded from the start of pretraining. Times are measured on a single NVIDIA L40. $\dagger$: The training follows the training recipe and data of LLaVA-1.5 due to the unavailability of a reproducible recipe of LLaVA-NeXT.}
\label{fig:result}
\end{figure*}

\begin{table*}[!t]
\vskip 0.05in
\caption{\textbf{Comparison of different VTP methods when applied in inference and training, respectively.} The pruning rate $p$ is uniformly set to 90$\%$ for fair comparison, whenever pruning is involved. Best and the second-best results are in \textbf{bold} and \underline{underlined}.}
\label{tab:different_vtp}
\centering
\resizebox{\textwidth}{!}{
\begin{tabular}{lcccccccccc}
\toprule
Methods & VQA\textsuperscript{v2} & GQA & SQA & VQA\textsuperscript{Text} & POPE & MME & MMB & MMB\textsuperscript{CN} & SEED & Rel. \\ \midrule
\rowcolor[HTML]{F2F2F2} 
LLaVA-1.5 & 79.16 & 62.82 & 68.62 & 57.73 & 86.19 & 1489.32 & 65.21 & 57.39 & 61.43 & 100.00$\%$ \\ \midrule
infer w/ FasterVLM & 74.12 & 54.52 & 68.57 & 54.86 & 72.34 & 1252.08 & 62.89 & 55.50 & 55.11 & 91.80$\%$ \\
infer w/ DivPrune & 74.33 & 57.37 & 69.21 & 54.32 & 80.43 & 1360.98 & 61.51 & 53.78 & 57.07 & \underline{93.98$\%$} \\
infer w/ CDPruner & 76.23 & 58.67 & 68.27 & 54.74 & 85.69 & 1369.69 & 63.06 & 53.09 & 58.45 & \textbf{95.53$\%$} \\ \midrule
\rowcolor[HTML]{E3F2D9} 
DualSpeed (FasterVLM) & 77.75 & 60.87 & 68.42 & 56.82 & 85.07 & 1457.33 & 64.35 & 54.98 & 58.79 & 97.78$\%$ \\
\rowcolor[HTML]{E3F2D9} 
DualSpeed (DivPrune) & 78.20 & 62.04 & 69.66 & 57.42 & 86.57 & 1490.45 & 65.98 & 56.87 & 59.72 & \textbf{99.61$\%$} \\
\rowcolor[HTML]{E3F2D9} 
DualSpeed (CDPruner) & 78.21 & 61.64 & 69.51 & 57.55 & 86.42 & 1482.38 & 65.29 & 57.22 & 59.90 & \underline{99.45$\%$} \\ \bottomrule
\end{tabular}
}
\vskip -0.1in
\end{table*}

\subsection{Experimental Setup}
\textbf{Configurations.}
We strictly follow the experimental settings of the widely used LLaVA-1.5~\cite{llava_1.5}, including models, training recipe, data, and hyperparameters.
The training process comprises two phases: in the pretraining phase, only the projector is trained, and in the Supervised Fine-Tuning (SFT) phase, both the projector and LLM are trained.
We adopt LoRA~\cite{lora} in the SFT phase for efficient experimentation while strictly keeping all comparisons fair.
According to its official results, training LLaVA-1.5 with LoRA achieves nearly equivalent performance to full-parameter tuning.
For training datasets, the pretraining phase utilizes LLaVA-Pretrain-558K~\cite{llava,llava_1.5} caption dataset, and the SFT phase utilizes LLaVA-665K~\cite{llava_1.5} instruction tuning dataset.
More details are in Appendix~\ref{app_sec:detailed_setup}.

\textbf{\model Implementation Details.}
For the choice of the plugable VTP, we adopt a simple yet effective method, DivPrune~\cite{divprune}, by default (different VTP options are compared in Section~\ref{sec:different_vtp}).
For hyperparameters, we fix the length $l$ of mode isolator to 4 (ablated in Section~\ref{sec:ablation}), and set the distillation temperature $\tau$ to 1 without hyperparameter tuning.
Unless otherwise specified, \model adopts $p=90\%$ and $r=10\%$ for the pruning ratio and the slow-mode activation probability, respectively (ablated in Section~\ref{sec:trade_off}).

\textbf{Compared Methods.}
We compare our \model to two methods: the baseline and the NaivePrune.
\textbf{\textit{Baseline}} refers to the model trained with its original training recipe.
\textbf{\textit{NaivePrune}} refers to directly applying VTP at training time.

\textbf{Evaluation.}
We comprehensively evaluate our method on 9 representative visual understanding benchmarks, including VQAv2~\cite{vqav2}, GQA~\cite{gqa}, SQA~\cite{sqa}, TextVQA~\cite{textvqa}, POPE~\cite{pope}, MME~\cite{mme}, MMBench~\cite{mmbench}, MMBench-CN~\cite{mmbench}, and SEED-Bench~\cite{seed_bench}.
Evaluations on these benchmarks follow their default settings and evaluation metrics (detailed in Appendix~\ref{app_sec:detailed_setup}).
We use relative performance (Rel.) to measure the performance retention of evaluated methods, which is computed by averaging the performance ratios of the evaluated model to the baseline across all benchmarks.
\textbf{Unless otherwise specified, all trained models are evaluated without applying pruning at inference time.} Since the research goal is to enable the model to infer on the full sequence to achieve uncompromised performance, as discussed in Section~\ref{sec:intro} and Figure~\ref{fig:close_gap}.

\subsection{Main Results}
In this subsection, we mainly seek the answers to the following research questions:
\begin{tcolorbox}
\textbf{Q1}: How much speedup and performance retention can \model achieve?
\tcbline
\textbf{Q2}: Do better VTP methods for inference achieve better performance when applied in training?
\tcbline
\textbf{Q3}: How significant is the training-inference gap?
\tcbline
\textbf{Q4}: To what extent are visual tokens redundant during training?
\end{tcolorbox}

\subsubsection{Comparison to Baselines}
\textbf{Results on LLaVA-1.5.}
We apply the proposed \model to accelerate the training of LLaVA-1.5-7B~\cite{llava_1.5}.
We compare the training time and performance of \model against the baseline, and the results are presented in Figure~\ref{fig:result_llava}.
It is observed that \model achieves an overall training speedup of 2.1$\times$ compared to the baseline while retaining 99.61$\%$ final performance.
Specifically, it yields a speedup of 5.8$\times$ in the pretraining phase, which is significantly higher than the overall speedup.
This is mainly attributed to the lower proportion of visual tokens in the SFT phase of LLaVA-1.5, where the training samples contain longer text queries and answers, thus reducing the contribution of VTP to the overall training acceleration (yet still achieving a significant overall speedup).
Notably, in the early stage of SFT phase, \model undergoes a rapid performance ascent, in contrast to the steady trend of performance improvement observed in the baseline.
We conjecture the reason is that the pruned visual sequence has a higher signal-to-noise ratio, which reduces the training difficulty in the early stage.

\textbf{Results on LLaVA-NeXT.}
We further explore the robustness of \model across models and its potential under higher visual resolutions by employing it on LLaVA-NeXT-7B~\cite{llava_next}, which converts each image into 2880 visual tokens.
The training recipe and data follow the same configuration of LLaVA-1.5\footnote{Due to the unavailability of a reproducible recipe of LLaVA-NeXT.}.
As shown in Figure~\ref{fig:result_llava_next}, it achieves an overall training speedup of 4.0$\times$ compared to the baseline while retaining 99.04$\%$ final performance.
Compared to LLaVA-1.5, higher-resolution models have a larger proportion of visual tokens, amplifying the speedup from VTP.
The consistent performance retention shows the robustness of \model across models, while the significantly higher speedup underscores its potential for high-resolution training.
Besides, \model undergoes a rapid performance ascent in the early stage of SFT phase, the same as observed in the experiment on LLaVA-1.5.

In summary, the above experiments across different MLLMs and resolutions demonstrate that:
\begin{tcolorbox}[title={Answer to Q1:}]
\model can achieve $\geq2\times$ training speedup, with $\geq99\%$ performance retention across models and visual resolutions.
\end{tcolorbox}

\begin{table*}[!t]
\caption{\textbf{Comparison of different methods evaluated under normal and pruning inference.} w/ mode isolator: pruning inference is performed with the trained mode isolator employed. The pruning rate $p$ is uniformly set to 90$\%$ for fair comparison, whenever pruning is involved. Best results in normal and pruning inference are in \textbf{bold}.}
\label{tab:train_infer_gap}
\centering
\resizebox{\textwidth}{!}{
\begin{tabular}{lcccccccccc}
\toprule
Methods & VQA\textsuperscript{v2} & GQA & SQA & VQA\textsuperscript{Text} & POPE & MME & MMB & MMB\textsuperscript{CN} & SEED & Rel. \\ \midrule
\rowcolor[HTML]{F2F2F2} 
LLaVA-1.5 & 79.16 & 62.82 & 68.62 & 57.73 & 86.19 & 1489.32 & 65.21 & 57.39 & 61.43 & 100.00$\%$ \\
+infer w/ prune & 74.33 & 57.37 & 69.21 & 54.32 & 80.43 & 1360.98 & 61.51 & 53.78 & 57.07 & 93.98$\%$ \\ \midrule
NaivePrune & 77.12 & 60.88 & 67.13 & 55.37 & 84.67 & 1385.28 & 63.06 & 51.72 & 59.49 & 95.89$\%$ \\
+infer w/ prune & 76.96 & 60.69 & 69.11 & 56.97 & 85.24 & 1404.05 & 65.03 & 56.96 & 58.53 & 97.85$\%$ \\ \midrule
\rowcolor[HTML]{E3F2D9} 
DualSpeed & 78.20 & 62.04 & 69.66 & 57.42 & 86.57 & 1490.45 & 65.98 & 56.87 & 59.72 & \textbf{99.61$\%$} \\
\rowcolor[HTML]{E3F2D9} 
+infer w/ prune & 76.95 & 61.18 & 69.26 & 56.27 & 85.78 & 1416.92 & 65.38 & 57.13 & 58.72 & \textbf{98.12$\%$} \\
\rowcolor[HTML]{E3F2D9} 
+w/ mode isolator & 77.12 & 61.77 & 69.01 & 55.61 & 86.15 & 1427.90 & 65.03 & 56.87 & 58.73 & 98.10$\%$ \\ \bottomrule
\end{tabular}
}
\vskip -0.1in
\end{table*}

\begin{figure*}[!t]
\vskip 0.2in
\centering
\begin{subfigure}[!t]{0.95\columnwidth}
\includegraphics[width=\columnwidth]{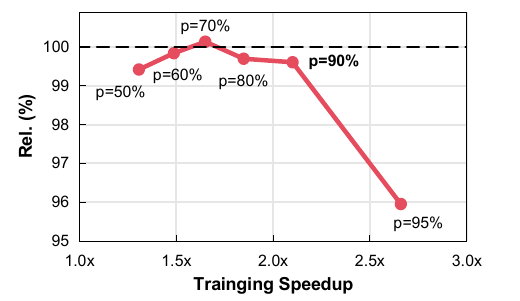}
\caption{Different values of $p$.}
\label{fig:trade_off_p}
\end{subfigure}
\begin{subfigure}[!t]{0.95\columnwidth}
\includegraphics[width=\columnwidth]{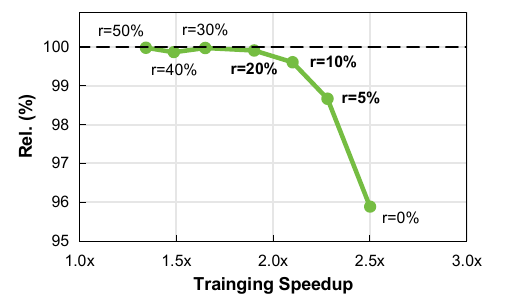}
\caption{Different values of $r$.}
\label{fig:trade_off_r}
\end{subfigure}
\caption{\textbf{Different speed-performance trade-offs under different $p$ and $r$.} We evaluate different $p=\{50,60,70,80,90,95\}\%$ by fixing $r$ to 10$\%$, and conversely, evaluate different $r=\{50,40,30,20,10,5,0\}\%$ by fixing $p$ to 90$\%$. When $r=0\%$, it degenerates to NaivePrune. Speedups are measured on a single NVIDIA L40.}
\label{fig:trade_off}
\end{figure*}

\subsubsection{Results for Different VTP Methods}\label{sec:different_vtp}
We compare the performance of \model when adopting different VTP methods, and the original performance of these VTP methods when applied in inference.
Specifically, we select three representative VTP methods: FasterVLM~\cite{fastervlm} (attention-based), DivPrune~\cite{divprune} (diversity-based), and CDPruner~\cite{cdpruner} (conditional-diversity-based).
Results are shown in Table~\ref{tab:different_vtp}.
When applied in inference, FasterVLM significantly lags behind the other two methods, and correspondingly, it also yields the lowest performance when applied in training.
Meanwhile, CDPruner achieves the best performance in inference, and yields near-best results (99.45$\%$ compares to the best 99.61$\%$ of DivPrune) when applied in training.
In short, at inference time: FasterVLM $\ll$ DivPrune $<$ CDPruner, and at training time: FasterVLM $\ll$ DivPrune $\approx$ CDPruner.
The relative performance relations among different VTP methods are not exactly consistent between inference and training, yet a general positive correlation is observed.
We conjecture that this slight discrepancy stems from their varying degrees of alignment with the training dynamics, \eg, CDPruner utilizes the textual queries to retain more answer-relevant visual tokens, which may damage the robustness of the model to handle answer-irrelevant visual tokens during inference.
Nonetheless, we can draw a general conclusion:
\begin{tcolorbox}[title={Answer to Q2:}]
Better VTP methods for inference \textit{tend to} yield better results when applied in training.
\end{tcolorbox}
It is worth clarifying that when \model adopts different VTP methods, the performance difference does not indicate poor robustness of it, \textbf{but rather reflects the inherent performance differences among distinct VTP methods}.

\begin{table*}[!t]
\caption{\textbf{Ablation of the key design choices.} Different variants are created by progressively adding each design choice on top of the NaivePrune. Speedups are measured on a single NVIDIA L40.}
\label{tab:ablate_design}
\centering
\resizebox{\textwidth}{!}{
\begin{tabular}{lccccccccccc}
\toprule
Methods & \begin{tabular}[c]{@{}c@{}}Training\\ Speedup\end{tabular} & VQA\textsuperscript{v2} & GQA & SQA & VQA\textsuperscript{Text} & POPE & MME & MMB & MMB\textsuperscript{CN} & SEED & Rel. \\ \midrule
\rowcolor[HTML]{F2F2F2} 
LLaVA-1.5 & 1.0$\times$ & 79.16 & 62.82 & 68.62 & 57.73 & 86.19 & 1489.32 & 65.21 & 57.39 & 61.43 & 100.00$\%$ \\ \midrule
NaivePrune & 2.5$\times$ & 77.12 & 60.88 & 67.13 & 55.37 & 84.67 & 1385.28 & 63.06 & 51.72 & 59.49 & 95.89$\%$ \\
+Fast-slow mode & 2.2$\times$ & 78.20 & 62.04 & 68.32 & 57.09 & 86.41 & 1443.98 & 65.03 & 57.65 & 59.29 & 98.88$\%$ \\
+Mode isolator & 2.2$\times$ & 78.26 & 62.06 & 69.06 & 57.17 & 86.18 & 1472.84 & 65.84 & 56.96 & 59.89 & 99.32$\%$ \\
+Self-distillation & 2.1$\times$ & 78.20 & 62.02 & 69.66 & 57.42 & 86.57 & 1490.45 & 65.98 & 56.87 & 59.72 & \textbf{99.61$\%$} \\ \bottomrule
\end{tabular}
}
\vskip -0.1in
\end{table*}

\subsubsection{Measure the Training-Inference Gap}\label{sec:train_infer_gap}
We compare the performance of models trained with the baseline, NaivePrune, and \model, under normal inference and pruning inference settings, as shown in Table~\ref{tab:train_infer_gap}.
Firstly, NaivePrune exhibits significantly lower performance than the normal inference baseline, whether it is 95.89$\%$ for normal inference or 97.85$\%$ for pruning inference, compared to the 100$\%$.
Generally, a model is expected to have lower performance in pruning inference than in normal inference.
However, NaivePrune achieves relatively higher performance in pruning inference (+1.96$\%$), since its training-inference is consistent at this point, \ie, both on pruned visual sequences.
Therefore, it implies that \textbf{the poor performance of pruning-trained models in normal inference is caused by training-inference mismatch}.
In contrast, \model mitigates this mismatch, thus bringing 3.72$\%$ performance improvement in normal inference, compared to NaivePrune.
Therefore, it is concluded that:
\begin{tcolorbox}[title={Answer to Q3:}]
The training-inference gap that \model can close is around 3.72$\%$.
\end{tcolorbox}
By closing this gap, \model achieves a comparable performance to the baseline (only -0.39$\%$).
As an additional advantage, it outperforms the baseline in pruning inference (+4.14$\%$), benefiting from its pruning-aware fast-mode training.
In other words, the model trained with \model is naturally compatible with both normal inference and pruning inference.
Especially, employing the trained mode isolator in pruning inference does not significantly affect the performance (98.10$\%$ compares to the original 98.12$\%$), despite the fact that it helps the training process, as ablated in Section~\ref{sec:ablation}.
We conjecture that it is analogous to the scaffolding theory that the mode isolator serves a temporary auxiliary role during the learning process and can eventually be removed in inference as the model's capabilities advance.

\subsubsection{Speed-Performance Trade-Off}\label{sec:trade_off}
The training speedup of \model depends on two hyperparameters: pruning ratio $p$ and the slow-mode activation probability $r$.
We first explore the trade-off between training speed and model performance under different values of $p$, as shown in Figure~\ref{fig:trade_off_p}.
It is observed that as the pruning ratio $p$ increases, training speed continuously improves while model performance remains nearly unchanged, until $p$ exceeds an extreme value of 90$\%$.
When $p$ increases from 90$\%$ to 95$\%$, the performance drops sharply by over 3$\%$.
Therefore, it comes to a conclusion that:
\begin{tcolorbox}[title={Answer to Q4:}]
Roughly up to 90$\%$ visual tokens are redundant for MLLM training.
\end{tcolorbox}
Similarly, we investigate the trade-off under different values of $r$, as shown in Figure~\ref{fig:trade_off_r}.
It is observed that the performance does not decline when $r$ is higher than 20$\%$, but drops sharply when $r$ exceeds 10$\%$.
The above results demonstrate the robustness of \model to $p$ and $r$.
In summary, when $p\approx 90\%$ and $r\approx 10\%$, the trade-off lies on the Pareto frontier, achieving maximum training speedup while preserving performance.

\subsection{Ablation Study}\label{sec:ablation}
\textbf{Ablate Design Choices.}
We ablate the key design choices of \model by progressively adding each design choice on top of the NaivePrune, to create different variants.
Ablation results are shown in Table~\ref{tab:ablate_design}.
There are two key observations:
\textit{(1) Fast-slow mode achieves a superior speed-performance trade-off.}
The design of fast-slow mode significantly improves the performance retention from 95.89$\%$ to 98.88$\%$, at the cost of merely 0.3$\times$ training speedup decline.
\textit{(2) Further improvements from mode isolator and self-distillation.}
Mode isolator improves the performance retention by 0.44$\%$, and self-distillation further improves 0.29$\%$, both with negligible impacts on the speedup.
Altogether, \model achieves 3.72$\%$ performance improvements over the 95.89$\%$ of the baseline NaivePrune, resulting in a 99.61$\%$ nearly lossless performance.

\begin{figure}[!t]
\vskip 0.2in
\centering
\includegraphics[width=0.9\columnwidth]{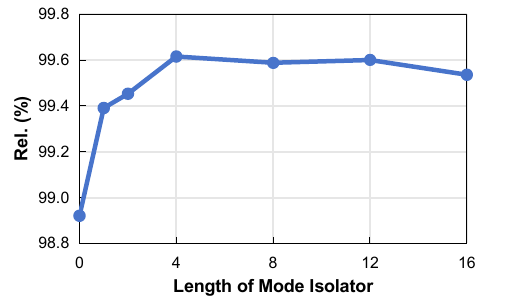}
\caption{\textbf{Ablation of the length $l$ of mode isolator.} We evaluate different $l=\{0,1,2,4,8,12,16\}$. A length of 0 indicates that the mode isolator is not employed.}
\label{fig:ablate_l}
\end{figure}

\textbf{Ablate Mode Isolator Length.}
We ablate the length of the proposed mode isolator as in Figure~\ref{fig:ablate_l}.
The performance improves when employing the mode isolator with a length of at least 1.
As the length $l$ increases, the performance first improves, and then saturates when $l$ is larger than 4.
Therefore, we set $l$ to 4 in all our experiments.

%% file: sections/appendix.tex
\section{Analysis of Training-Inference Mismatch}

\begin{table}[!t]
\caption{\textbf{Apply VTP in different training phases.} Here, we adopt DivPrune for VTP, with a pruning ratio of 90$\%$.}
\label{tab:vtp_different_phases}
\centering
\resizebox{\textwidth}{!}{
\begin{tabular}{cccccccccccc}
\toprule
\begin{tabular}[c]{@{}c@{}}Apply VTP in\\ Pretraining\end{tabular} & \begin{tabular}[c]{@{}c@{}}Apply VTP in\\ SFT\end{tabular} & VQA\textsuperscript{v2} & GQA & SQA & VQA\textsuperscript{Text} & POPE & MME & MMB & MMB\textsuperscript{CN} & SEED & Rel. \\ \midrule
\textcolor[HTML]{e54c5e}{\ding{56}} & \textcolor[HTML]{e54c5e}{\ding{56}} & 79.16 & 62.82 & 68.62 & 57.73 & 86.19 & 1489.32 & 65.21 & 57.39 & 61.43 & 100.00$\%$ \\
\textcolor[HTML]{75bd42}{\ding{52}} & \textcolor[HTML]{e54c5e}{\ding{56}} & 79.32 & 62.88 & 68.70 & 56.83 & 86.64 & 1476.90 & 66.10 & 58.23 & 61.27 & 100.12$\%$ \\
\textcolor[HTML]{e54c5e}{\ding{56}} & \textcolor[HTML]{75bd42}{\ding{52}} & 77.07 & 60.27 & 67.54 & 55.85 & 84.31 & 1307.68 & 64.67 & 52.33 & 60.21 & 95.83$\%$ \\
\textcolor[HTML]{75bd42}{\ding{52}} & \textcolor[HTML]{75bd42}{\ding{52}} & 77.12 & 60.88 & 67.13 & 55.37 & 84.67 & 1385.28 & 63.06 & 51.72 & 59.49 & 95.89$\%$ \\ \bottomrule
\end{tabular}
}
\vskip -0.1in
\end{table}

\subsection{Theoretical Analysis of the Root Cause}\label{app_sec:cause_of_mismatch}
Training-inference mismatch is the core challenge of applying VTP to MLLM training.
Its essential cause lies in the discrepancies between the pruned sequences used in training and the full sequences used in normal inference.
The discrepancies are in three key aspects: sequence length, information density, and spatial structure.

\textbf{Sequence Length Mismatch.}
The model optimizes on short sequences during training, forming biased attention weights and simplified context interaction logic, which disables reasonable attention allocation and new token dependency capture for extended sequences in inference, causing suboptimal feature fusion and prediction errors.

\textbf{Information Density Mismatch.}
VTP retains high-information tokens while eliminating low-information redundancies.
Training relies on high signal-to-noise ratio inputs, focusing only on key visual-textual correlations.
In inference, the model is incompatible with the full sequence’s low-density feature distributions, leading to feature shift and prediction deviations.

\textbf{Spatial Structure Mismatch.}
Visual tokens carry not only semantic information but also spatial positional information (\eg, object relative positions, scene layout).
Pruning disrupts the original spatial topology, leading the model to learn biased representations.
In inference, the restored full-structure sequence confuses the model’s spatial perception, impairing its overall performance.

In summary, the three discrepancies collectively lead to the training-inference mismatch of VTP-based training.
The dual-mode design of \model fundamentally addresses this problem by enabling the model to learn both pruned and full sequences simultaneously.

\subsection{Analysis of Which Part of Model Suffers from Mismatch}\label{app_sec:which_part_mismatch}
\textbf{Theoretically}, the aforementioned discrepancies mainly affect the self-attention mechanism of the LLM, rather than the multimodal projector.
This is because the learning goal of the multimodal projector is to align feature spaces, rather than capturing sequence patterns.

\textbf{Empirically}, to verify this, we conduct a controlled experiment.
As shown in Figure~\ref{tab:vtp_different_phases}, when VTP is solely employed in the pretraining phase, there is no degradation in the final performance (100.12$\%$ compared to the original 100.00$\%$), and there is even a slight performance improvement.
We conjecture that VTP eliminates low-contribution visual tokens, allowing the projector to focus on learning mappings of key visual features.
However, when VTP is employed in the SFT phase, the performance degradation is the same as when it is employed in both phases (95.83$\%$ compared to 95.89$\%$).

\begin{table}[!t]
\caption{\textbf{Training recipe for both LLaVA-1.5 and LLaVA-NeXT.}}
\label{tab:training_recipe}
\centering
\begin{tabular}{lcc}
\toprule
Configurations & Pretraining & SFT \\ \midrule
Trainable Components & projector & projector and LLM \\
Batch Size & 256 & 128 \\
LoRA Rank & - & 128 \\
LoRA Alpha & - & 256 \\
Optimizer & \multicolumn{2}{c}{AdamW} \\
Learning Rate of Projector & 1e-3 & 2e-5 \\
Learning Rate of LLM & - & 2e-4 \\
Learning Rate Scheduler & \multicolumn{2}{c}{cosine decay} \\
Warmup Ratio & \multicolumn{2}{c}{0.03} \\
Weight Decay & \multicolumn{2}{c}{0} \\
Epoch & \multicolumn{2}{c}{1} \\ \midrule
Data & \begin{tabular}[c]{@{}c@{}}LLaVA-Pretrain-558K\\ \cite{llava,llava_1.5}\end{tabular} & \begin{tabular}[c]{@{}c@{}}LLaVA-665K\\ \cite{llava_1.5}\end{tabular} \\ \bottomrule
\end{tabular}
\vskip -0.1in
\end{table}

\section{Detailed Experimental Setup}\label{app_sec:detailed_setup}

\textbf{Training Recipe.}
We configure the training recipe strictly the same as LLaVA-1.5~\cite{llava_1.5}, as listed in Table~\ref{tab:training_recipe}.
For training that involves updating the LLM, we utilize  LoRA~\cite{lora} for efficient experimentation while strictly keeping all comparisons fair.
It is applied across all linear modules within the LLM.
According to its official results, training LLaVA-1.5 with LoRA achieves nearly equivalent performance to full-parameter tuning.

\textbf{Model Implementations.}
For LLaVA-1.5-7B and LLaVA-NeXT-7B, they adopt CLIP-ViT-L/14~\cite{clip_vit_l_14} as the visual encoder, a two-layer MLP with GELU activation function as the projector, and Vicuna-v1.5-7B~\cite{vicuna} as the LLM.

\textbf{Evaluation Details.}
The evaluations on certain benchmarks are specified following their common practice.
SQA~\cite{sqa} evaluates on its image split.
POPE~\cite{pope} evaluates on all three splits and reports the averaged F1 score.
MME~\cite{mme} evaluates on its Perception split.
SEED-Bench~\cite{seed_bench} evaluates on both image and video splits, and reports the averaged score.
The evaluation of training time or speedup is measured on a single NVIDIA L40 48G.